\documentclass[a4paper]{article}
\usepackage{INTERSPEECH_v2}
\usepackage{url,bm,amssymb,amsmath,graphicx}
\usepackage{appendix}
\usepackage{color}
\usepackage{placeins}
\usepackage{balance}

\usepackage{float}
\usepackage{multirow}

\title{End-to-end Conversation Modeling Track in DSTC6}
\name{Chiori Hori, Takaaki Hori}

\address{Mitsubishi Electric Research Laboratories\\
201 Broadway, Cambridge, MA 02139, USA}
\email {\{chori, thori\}@merl.com}

\begin{document}
\maketitle

\begin{abstract}
  This article presents a brief overview of the end-to-end conversation modeling track of the 6th dialog system technology challenges (DSTC6).
  The task was to develop　a fully data-driven dialog system using a customer service conversations from Twitter business feeds, and invited participants who work on this challenge track. 
  In this overview, we describe the task design and data sets, and review the submitted systems and applied techniques for conversation modeling. 
  We received 19 system outputs from six teams, and evaluated them based on several objective measures and a human-rating based subjective measure. 
  Finally, we discuss technical achievements and remaining problems related to this challenge. 
\end{abstract}

\section{Introduction}
End-to-end training of neural networks is a promising approach to automatic construction of dialog systems using a human-to-human dialog corpus. Recently, Vinyals et al. tested neural conversation models using OpenSubtitles \cite{vinyals2015neural}. Lowe et al. released the Ubuntu Dialogue Corpus \cite{lowe2015ubuntu} for research in unstructured multi-turn dialogue systems. Furthermore, the approach has been extended to accomplish task oriented dialogs to provide information properly with natural conversation. For example, Ghazvininejad et al. proposed a knowledge grounded neural conversation model \cite{ghazvininejad2017knowledge}, where the research is aiming at combining conversational dialogs with task-oriented knowledge using unstructured data such as Twitter data for conversation and Foursquare data for external knowledge.  However, the task is still limited to a restaurant information service, and has not yet been tested with a wide variety of dialog tasks. In addition, it is still unclear how to create intelligent dialog systems that can respond like a human agent.

In consideration of these problems, we proposed a challenge track to the 6th dialog system technology challenges (DSTC6) \cite{dstc6}.
The focus of the challenge track is to train end-to-end conversation models from human-to-human conversation in order to accomplish end-to-end dialog tasks for a customer service. 
The dialog system plays the role of a human agent and generates natural and informative sentences in response to user’s questions or comments given a dialog context. 
 
\section{Tasks}
\begin{figure*}[tbh]
\centering
\centerline{\includegraphics[width=12.0cm]{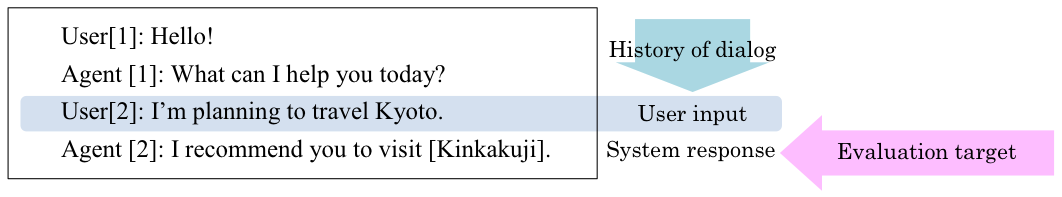}}
\caption{Sentence generation and evaluation in the end-to-end conversation modeling track.}
\label{fig:task}
\end{figure*}

In this challenge track, a system has to generate sentence(s) in response to a user input in a given dialog context, where it can use external knowledge from public data, e.g. web data. The quality of the automatically generated sentences is evaluated with objective and subjective measures to judge whether or not the generated sentences are natural and informative for the user (see Fig. \ref{fig:task}).

This track consists of two tasks, a main task and a pilot task:
\begin{description} 
\item[(1) Main task:] Customer service dialog using Twitter data (mandatory)
\begin{description} 
  \item[Task A:] Full or part of the training data will be used to train conversation models. 
  \item[Task B:] Any open data, e.g. from web, are available as external knowledge to generate informative sentences. But they should not overlap with the training, validation and test data provided by organizers.
\end{description}
Challenge attendees can select either A or B, or both. The tools to download Twitter data and extract the dialog text were provided to all attendees at the challenge track in DSTC6 \cite{E2EConversation_github}. The attendees needed to collect the data by themselves. Data collected before Sep. 1st, 2017 was available as trial data, and the official training, development and test data were collected from Sep. 7th to 18th, 2017. The dialogs were used for the test set was not disclosed until Sep. 25th.
 
\item[(2) Pilot task:] Movie scenario dialog using OpenSubtitles data (optional)\\
     The OpenSubtitles are basically not task-oriented dialogs, but the naturalness and correctness of system responses will be evaluated.
\end{description} 

\section{Data Collection}
\subsection{Twitter data}
In the Twitter task, we used dialog data collected from multiple Twitter accounts for customer service. Each dialogue consisted of real tweets between a customer and an agent. A customer usually asked a question or complained something about a product or a service of the company, and an agent responded to the customer accordingly. In this challenge, each participant is supposed to develop a dialog system that mimics agents’ behaviors. The system will be evaluated based on the quality of generated sentences in response to customers’ tweets.
For the challenge, we provided a data collection tool to all participants so that they could collect the data by themselves because Twitter does not allow distribution of Twitter data by a third party
In this task, it is assumed that each participant continued to collect the data from specific accounts in the challenge period. To acquire a large amount of data, the data collection needed to be done repeatedly, e.g. by running the script once a day, because the amount of data we can download is limited and older tweets cannot be accessed after they expire. 
At a certain point of time, we provided an additional tool to extract subsets of collected data for training, development (validation), and evaluation so that all the participants were able to use the same data for the challenge. Until the official data sets were fixed, trial data sets were available to develop dialog systems, which were selected from the data collected by each participant. But once the official data sets were determined, the system needed to be trained from scratch only using the official data sets.

Challenge attendees need to use a common data collection tool ``\texttt{collect\_twitter\_dialogs}'' included in the provided package. Necessary steps are written in ``\texttt{collect\_twitter\_dialogs/README.md}''
The trial data sets can be extracted from downloaded twitter dialogs using a data extraction script: 
``\texttt{make\_trial\_data.sh}'' in ``\texttt{tasks/twitter}''. 

The official data are collected through the period of Sep. 7th to 18th in 2017,
using the data collection tool. The official training, development and test sets can be
extracted using a data extraction script: ``\texttt{make\_official\_data.sh}'' in ``\texttt{tasks/twitter}''.
Finally, the participants are supposed to collect the data sets summarized in Table \ref{tab:twitter_data}. 
You can find more information in "https://github.com/dialogtekgeek/DSTC6-End-to-End-Conversation-Modeling".

\begin{table}[t]
\centering
\caption{Twitter data.}
\label{tab:twitter_data}
\begin{tabular}{ll|ccc}
\hline
        & & training & development & test \\
        \hline
\#dialog & & 888,201    & 107,506  & 2,000 \\
\#turn   & & 2,157,389  & 262,228  & 5,266 \\
\#word   & & 40,073,697 & 4,900,743 & 99,389 \\
\hline
\end{tabular}
\end{table}

\subsection{OpenSubtitles data}
OpenSubtitles is a freely availale corpus of movie subtitles, which is available from a website\footnote{\url{http://www.opensubtitles.org/}}.  We used English subtitles in OpenSubtitles2016 corpus \cite{lison2016opensubtitles2016}\footnote{ \url{http://opus.lingfil.uu.se/OpenSubtitles2016.php}}.
The text data file can be downloaded from \url{http://opus.lingfil.uu.se/download.php?f=OpenSubtitles2016/en.tar.gz}.

The file size of ``\texttt{en.tar.gz}'' is approximately 18GB. 
The training, development and test sets can be extracted from the decompressed directory using a data extraction script: ``\texttt{make\_trial\_data.sh}'' in ``\texttt{tasks/opensubs}''. 
The official data sets can also be extracted using ``\texttt{make\_official\_data.sh}.''
You can download the tools from "https://github.com/dialogtekgeek/DSTC6-End-to-End-Conversation-Modeling".
The sizes of the official data sets are summarized in Table \ref{tab:opensubs_data}.
\begin{table}[t]
\centering
\caption{OpenSubtitles data.}
\label{tab:opensubs_data}
\begin{tabular}{ll|ccc}
\hline
        & & training & development & test \\
        \hline
\#dialog & & 31,073,509  & 310,865 & 2,000 \\
\#turn   & & 62,147,018  & 621,730 & 4,000 \\
\#word   & & 413,976,295 & 4,134,686 & 26,611 \\
\hline
\end{tabular}
\end{table}

\section{Text Preprocessing}
Twitter dialogs and OpenSubtitles contain a lot of noisy text with special expressions and symbols.  Therefore, text preprocessing is important to clean up and normalize the text. Moreover, all the participants need to use the same preprocessing at least for target references to assure fair comparisons between different systems in the Challenge.
 
\subsection{Twitter data}
Twitter data contains a lot of specific information such as Twitter account names, URLs, e-mail addresses, telephone/tracking numbers and hashtags. This kind of information is almost impossible to predict correctly unless we use a lot of training data obtained from the same site. To alleviate this difficulty, we substitute those strings with abstract symbols such as \texttt{<URL>}, \texttt{<E-MAIL>}, and \texttt{<NUMBERS>} using a set of regular expressions. In addition, since each tweet usually starts with a Twitter account name of the recipient, we removed the account name. But if such names appear within a sentence, we leave them because those names are a part of sentence. We leave hashtags as well for the same reason. We also substitute user names with \texttt{<USER>} e.g.
\begin{align}
    \texttt{hi~} & \texttt{John, can you send me a dm ?} \nonumber \\
    & \rightarrow ~~~ \texttt{hi <USER>, can you send me a dm ?} \nonumber
\end{align}
 
Since the user's name can be extracted from the attribute information of each tweet, we can replace it. Note that the text preprocessing is not perfect, and therefore there may remain original phrases, which are not replaced or removed successfully. The text also includes many abbreviations, e.g. ``pls hlp'', special symbols, e.g. ``\texttt{(-:}'' and wide characters ``$\copyright\heartsuit\clubsuit$'' including 4-byte Emojis. These are left unaltered. The wide characters are encoded with UTF-8 in the text file.
The preprocessing is performed by function ``\texttt{preprocess()}'' in python script ``\texttt{tasks/twitter/extract\_twitter\_dialogs.py}''.
The data format of the data sets are summarized in Appendix \ref{apndx:data_format}.
 
\subsection{OpenSubtitles data}
We basically follow the method in \cite{vinyals2015neural} to extract dialogs and preprocess the subtitle data. Two consecutive sentences are considered a dialog consisting of one question and one answer. The data format is the same as the Twitter data.

\section{Submitted Systems}
We received 19 sets of system outputs for the Twitter task, from six teams, and four system description papers were accepted \cite{wang2017sequence}\cite{long2017knowledge}\cite{bairong2017comparative}\cite{galley2017msrnlp}.  In this section, we summarize the techniques used in the systems, including the baseline system for the challenge track. There was no system submitted to the OpenSubtitles task, so we present the techniques and results only for the Twitter task.

\begin{table*}[t]
\centering
\caption{Submitted systems.}
\label{tab:systems}
\vskip -2mm
\begin{tabular}{l|l|l|l|c}
\hline
Team (Entry) & Model type& Objective function & Additional techniques & Paper \\
\hline
\hline
baseline     & LSTM   & Cross entropy &      \\
\hline
team\_1 (1)  &        &            & \\
team\_1 (2)  &         &           &  \\
\hline
team\_2 (1)  & BLSTM+LSTM & Adversarial & Example-based method & \\
team\_2 (2)  & LSTM       & Cross entropy & Example-based method & \\
team\_2 (3)  & BLSTM+LSTM & Adversarial+CosineSimilarity & Example-based method &  \cite{wang2017sequence}\\
team\_2 (4)  & BLSTM+LSTM & Adversarial &         &\\
team\_2 (5)  & BLSTM+LSTM+HRED & Cross entropy & MBR system combination &\\
\hline
team\_3 (1) & LSTM       & Cross entropy &      &\\
team\_3 (2) & 2LDTM + attention & Cross entropy on diversified data &      &  \\
team\_3 (3) & 2LSTM + attention & Cross entropy (tuned with trial data$^+$) &      & \cite{long2017knowledge}\\
team\_3 (4) & GWGM + attention & Cross entropy &      &\\
team\_3 (5) & SEARG            & Cross entropy & Knowledge enhanced model &    \\
\hline
team\_4 (1) & LSTM     & Cross entropy & Word embedding initialization &  \cite{bairong2017comparative}   \\
\hline
team\_5 (1) & LSTM     & MMI maxdiv &             &      \\
team\_5 (2) & LSTM     & MMI maxBLEU (tuned with MERT) &     &      \\
team\_5 (3) & LSTM     & MMI mixed (maxdiv + maxBLEU)  &             & \cite{galley2017msrnlp}     \\
team\_5 (4) & LSTM     & MMI uniform &             &      \\
team\_5 (5) & LSTM     & Cross entropy & Greedy search for decoding &      \\
\hline
team\_6 (1) &      &            &             &\\
\hline
\multicolumn{5}{r}{$^+$Trial data cannot be used for official evaluation. The results are not officially accepted.}
\end{tabular}
\end{table*}
The baseline system is based on an LSTM-based encoder decoder in \cite{E2EConversation_github}, but this is a simplified version of \cite{vinyals2015neural}, in which back-propagation is performed only up to the previous turn from the current turn, although the state information is taken over throughout the dialog.

Table \ref{tab:systems} shows the baseline and submitted systems with their brief specifications including model type, objective function, and additional techniques.
An empty specification means that the team did not submit any system description paper to the DSTC6 workshop.

Most systems employed a LSTM or BLSTM (2LSTM) encoder and a LSTM decoder. Some systems used a hierarchical encoder decoder (team\_2(5) and team\_3(5)) and attention-based decoder (team\_3(2,3,4)).
Several types of objective functions were applied for training the models, where cross entropy, adversarial method, cosine similarity, and maximum mutual information (MMI) were used solely or combined.
The objective functions except cross entropy were designed to increase the diversity of responses. 
This hopefully leaded more realistic and informative responses.

Furthermore, several additional techniques are introduced to improve the response quality. In \cite{wang2017sequence}, an example-based method is used to return real human responses if a similar context exists in the training corpus, and
minimum Bayes risk (MBR) decoding is used to improve objective scores.
The knowledge enhanced encoder decoder \cite{long2017knowledge} searches for relevant documents in the web using the keywords in the dialog context, and the relevant documents are used to enhance the decoder.  
In \cite{bairong2017comparative}, different types of word-embedding vectors are used for initialization of the models.

\section{Evaluation}
Challenge participants were allowed to  submit up to 5 sets of system outputs. 
The outputs were evaluated with objective measures such as BLEU and METEOR, and also evaluated by rating scores collected by humans using Amazon Mechanical Turk (AMT). The human evaluators rate the system responses in terms of naturalness, informativeness, and appropriateness.

\subsection{Objective evaluation}
For the challenge track, we used \texttt{nlg-eval}\footnote{\url{https://github.com/Maluuba/nlg-eval}} for objective evaluation of system outputs, which is a publicly available tool supporting various unsupervised automated metrics for natural language generation.
The supported metrics include word-overlap-based metrics such as BLEU, METEOR, ROUGE\_L, and CIDEr, and embedding-based metrics such as SkipThoughts Cosine Similarity, Embedding Average Cosine Similarity, Vector Extrema Cosine Similarity, and Greedy Matching Score.
Details of these metrics are described in \cite{sharma2017nlgeval}.

We prepared 10 more references for a ground truth of each response by humans to operate reliable objective evaluation. The references included a real human response in the Twitter dialog and 10 human-generated responses.
We asked 10 different Amazon Mechanical Turkers for each dialog to compose a sentence for the final response given the dialog context. 
We provided the real human response as an example and asked them to make their responses 
to be different from the example while keeping to the dialog topic. We also asked them not to copy and paste the example in their response.
When multiple references are available, \texttt{nlg-eval} computes the similarity between
the prediction and all the references one-by-one, and then selects the maximum value. 

\subsection{Subjective evaluation}
We collected human ratings for each system response using 5 point Likert Scale, where 10 different Turkers rated system responses given a dialog context.
We listed randomly 21 responses below the dialog context, which consists of 19 submitted outputs, a baseline output, and a human response for each dialog.

The Turkers rated each response by 5 level scores as
\begin{center}
\begin{tabular}{l|c}
\hline
Level & Score \\
\hline
Very good & 5\\
Good & 4 \\
Acceptable & 3 \\
Poor & 2 \\
Very poor & 1 \\
\hline
\end{tabular}
\end{center}
we instructed to the Turkers to consider naturalness, informativeness, and appropriateness of the response for the given context.
If there were identical responses in the list, we reduced them into one response
so that they were rated consistently.
The average score was computed for each system and reported in Table \ref{tab:all_results}.

\subsection{Results}
Table \ref{tab:all_results} shows evaluation results of 21 systems: the 19 submitted systems, the baseline and the reference. 
Systems are listed as team\_M(N), where M is the team index and N is an identifier for a particular system submitted by that team. 
``Ext. Data'' in the table denotes whether or not the system used external data for training and/or testing, where only team\_3(5) used external data (web data) for response generation.
\begin{table*}[p]
\centering
\caption{Evaluation results with objective measures based on 11 references and a subjective measure based on 5-level ratings}
\label{tab:all_results}
\vskip -3mm
\begin{tabular}{l|p{5mm}|cccccccc|c}
\hline
Team (Entry) & Ext. & BLEU4 & METEOR & ROUGE\_L & CIDEr & Skip & Embedding & Vector & Greedy & Human \\
            & Data     &        &         &         &       & Thought & Average & Extrema & Matching & Rating \\
\hline
\hline
baseline &  & 0.1619 & 0.2041 & 0.3598 & 0.0825 & 0.6380 & 0.9132 & 0.6073 & 0.7590 & 3.3638 \\
\hline
team\_1 (1)$^*$ & & 0.1598 & 0.2020 & 0.3608 & 0.0780 & 0.6451 & 0.9090 & 0.6039 & 0.7572 & 3.4415 \\
team\_1 (2)$^*$ & & 0.1623 & 0.2039 & 0.3567 & 0.0828 & 0.6386 & 0.9026 & 0.6071 & 0.7587 & 3.4297 \\
\hline
team\_2 (1) & & 0.1504 & 0.1826 & 0.3446 & 0.0803 & 0.6451 & 0.9070 & 0.5990 & 0.7534 & 3.4453 \\
team\_2 (2) & & 0.2118 & 0.2140 & 0.3953 & 0.1060 & {\bf 0.7075} & {\bf 0.9271} & 0.6371 & 0.7747 & 3.3894 \\
team\_2 (3) & & 0.1851 & 0.2040 & 0.3748 & 0.0965 & 0.6706 & 0.9116 & 0.6155 & 0.7613 & 3.4777 \\
team\_2 (4) & & 0.1532 & 0.1833 & 0.3469 & 0.0800 & 0.6463 & 0.9077 & 0.5999 & 0.7544 & 3.4381 \\
team\_2 (5) & & 0.2205 & {\bf 0.2210} & {\bf 0.4102} & {\bf 0.1279} & 0.6636 & 0.9251 & {\bf 0.6449} & {\bf 0.7802} & 3.4332 \\
\hline
team\_3 (1) & & 0.1602 & 0.2016 & 0.3606 & 0.0782 & 0.6474 & 0.9074 & 0.6031 & 0.7567 & 3.4503 \\
team\_3 (2) & & 0.1779 & 0.2085 & 0.3829 & 0.0978 & 0.6259 & 0.9201 & 0.6106 & 0.7683 & 3.5239 \\
team\_3 (3)$^{**}$ & & 0.1741 & 0.2024 & 0.3703 & 0.0994 & 0.6348 & 0.8985 & 0.6000 & 0.7573 & 3.5082 \\
team\_3 (4) & & 0.1342 & 0.1762 & 0.3366 & 0.0947 & 0.6127 & 0.8802 & 0.5913 & 0.7412 & 3.5107 \\
team\_3 (5) & ~~\checkmark & 0.1092 & 0.1731 & 0.3201 & 0.0702 & 0.6132 & 0.8977 & 0.5870 & 0.7420 & 3.3919 \\
\hline
team\_4 (1) & & 0.1716 & 0.2071 & 0.3671 & 0.0898 & 0.6529 & 0.9106 & 0.6079 & 0.7596 & 3.4431 \\
\hline
team\_5 (1) & & 0.1480 & 0.1813 & 0.3388 & 0.1025 & 0.6131 & 0.9087 & 0.5928 & 0.7433 & 3.5209 \\
team\_5 (2) & & 0.0991 & 0.1687 & 0.3146 & 0.0708 & 0.5952 & 0.8996 & 0.5675 & 0.7257 & 3.3053 \\
team\_5 (3) & & 0.1448 & 0.1839 & 0.3375 & 0.0940 & 0.6025 & 0.9083 & 0.5915 & 0.7433 & {\bf 3.5396} \\
team\_5 (4) & & 0.1261 & 0.1754 & 0.3310 & 0.0945 & 0.6151 & 0.8984 & 0.5814 & 0.7330 & 3.4545 \\
team\_5 (5) & & 0.1575 & 0.1918 & 0.3658 & 0.1112 & 0.6457 & 0.9076 & 0.6075 & 0.7528 & 3.5097 \\
\hline
team\_6 (1)$^*$ & & {\bf 0.2762} & 0.1656 & 0.3482 & 0.1235 & 0.6989 & 0.8018 & 0.5854 & 0.7316 & 2.9906 \\
\hline
reference & &  &  &  &  &  &  &  &  & 3.7245 \\
\hline
\multicolumn{11}{r}{$^*$Results are not officially accepted since any system description paper has not been submitted. }\\
\multicolumn{11}{r}{$^{**}$Results are not officially accepted since the system was tuned with the trial data \cite{long2017knowledge}.}
\end{tabular}
\vskip 2mm
\centering
\caption{Cross validation results of 11 references.}
\label{tab:cross-val}
\vskip -3mm
\begin{tabular}{l|cccccccc}
\hline
 & BLEU4 & METEOR & ROUGE\_L & CIDEr & Skip & Embedding & Vector & Greedy \\
 &        &         &         &       & Thought & Average & Extrema & Matching \\
\hline
Original & 0.5264 & 0.3885 & 0.6559 & 0.7566 & 0.7160 & 0.9483 & 0.7625 & 0.8679 \\
Ref (1) & 0.2758 & 0.2357 & 0.4525 & 0.3615 & 0.7091 & 0.9308 & 0.6643 & 0.7958 \\
Ref (2) & 0.2626 & 0.2313 & 0.4501 & 0.3477 & 0.7142 & 0.9300 & 0.6646 & 0.7951 \\
Ref (3) & 0.2651 & 0.2335 & 0.4499 & 0.3571 & 0.7064 & 0.9328 & 0.6626 & 0.7958 \\
Ref (4) & 0.2683 & 0.2358 & 0.4509 & 0.3622 & 0.7039 & 0.9313 & 0.6637 & 0.7951 \\
Ref (5) & 0.2682 & 0.2390 & 0.4464 & 0.3611 & 0.7104 & 0.9301 & 0.6632 & 0.7925 \\
Ref (6) & 0.2786 & 0.2323 & 0.4476 & 0.3577 & 0.7005 & 0.9291 & 0.6642 & 0.7925 \\
Ref (7) & 0.2729 & 0.2382 & 0.4523 & 0.3678 & 0.7049 & 0.9319 & 0.6687 & 0.7971 \\
Ref (8) & 0.2593 & 0.2256 & 0.4430 & 0.3488 & 0.7082 & 0.9306 & 0.6604 & 0.7921 \\
Ref (9) & 0.2529 & 0.2348 & 0.4436 & 0.3440 & 0.7202 & 0.9325 & 0.6621 & 0.7944 \\
Ref (10) & 0.2707 & 0.2364 & 0.4527 & 0.3750 & 0.7105 & 0.9333 & 0.6679 & 0.7968 \\
\hline
Average & 0.2910 & 0.2483 & 0.4677 & 0.3945 & 0.7095 & 0.9328 & 0.6731 & 0.8014 \\
\hline
\end{tabular}
\vskip -4mm
\end{table*}
Table \ref{tab:cross-val} shows cross-validation results of 11 ground truths generate by humans. Each manually generated sentence was evaluated by comparing with other 10 ground truths using leave-one-out method. 

In most objective measures, the system of team\_2(5) achieved highest scores, where the system employed MBR decoding for system combination.
This result indicates that explicit maximization of objective measures and the complementarity of multiple systems bring significant improvement for the objective measures.

On the subjective measure with human rating, the system of team\_5(3) achieved the best score (3.5396).
Although there were no big differences between the human rating scores,
we can see that techniques for improving human rating actually contributed to increase the rating scores. For example, adversarial training (team\_2(3)), use of diversified data (team\_3(2)), and MMI-based objective function with maximum diversity (team\_5(1),(3)) had some improvement compared to the other systems.
The automatic generation have not yet reached the score of the reference (3.7245), 
but some of them were better responses than the reference in terms of human rating. 
The conversation modeling techniques applied to this challenge track for Twitter dialogs are better than we expected.

\begin{figure*}[t]
\centering
\centerline{\includegraphics[width=13cm]{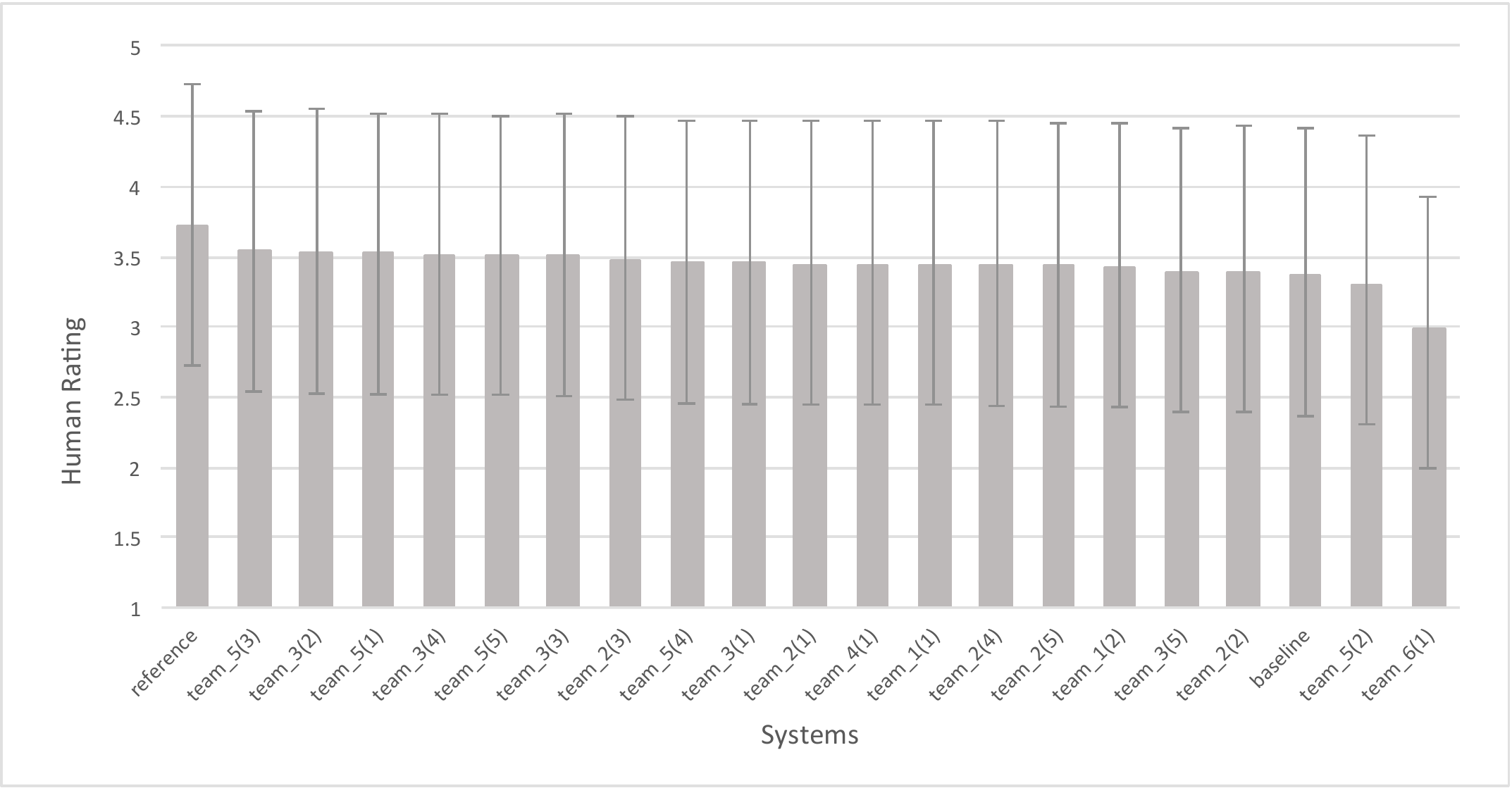}}
\vskip -3mm
\caption{Mean and standard deviation of human rating score.}
\label{fig:human-rating}
\centerline{\includegraphics[width=15cm]{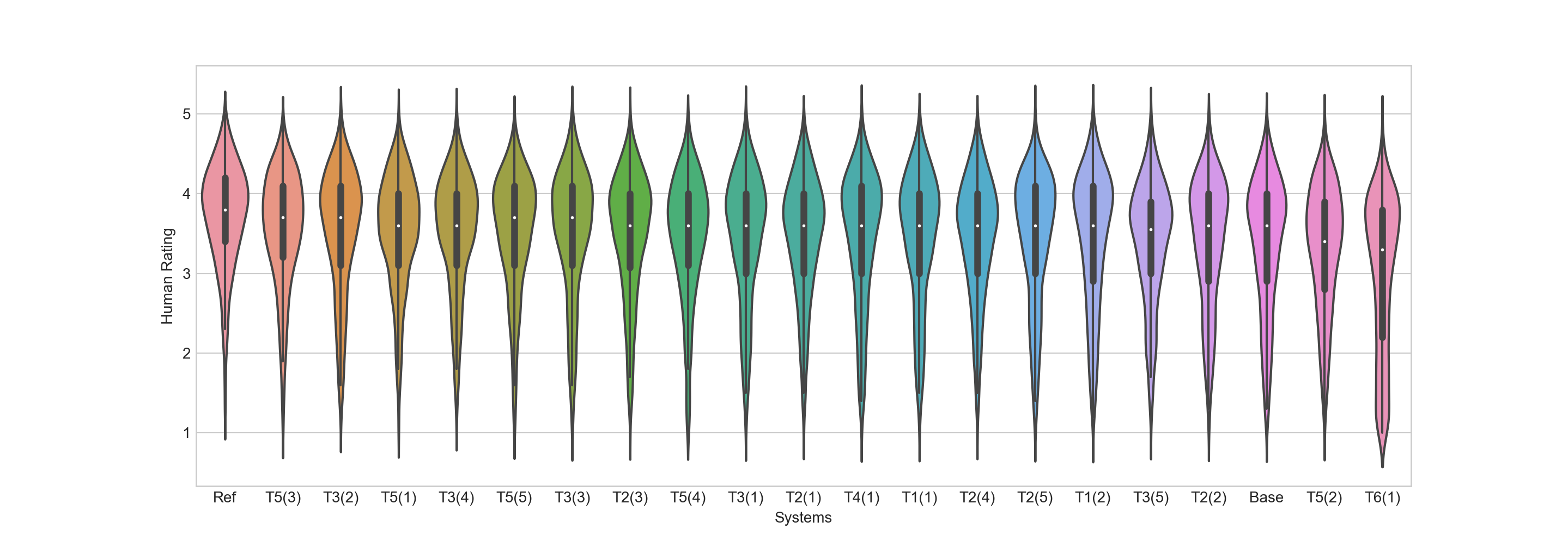}}
\vskip -3mm
\caption{Distribution of human scores averaged sentence by sentence.}
\label{fig:rating-violin1}
\centerline{\includegraphics[width=15cm]{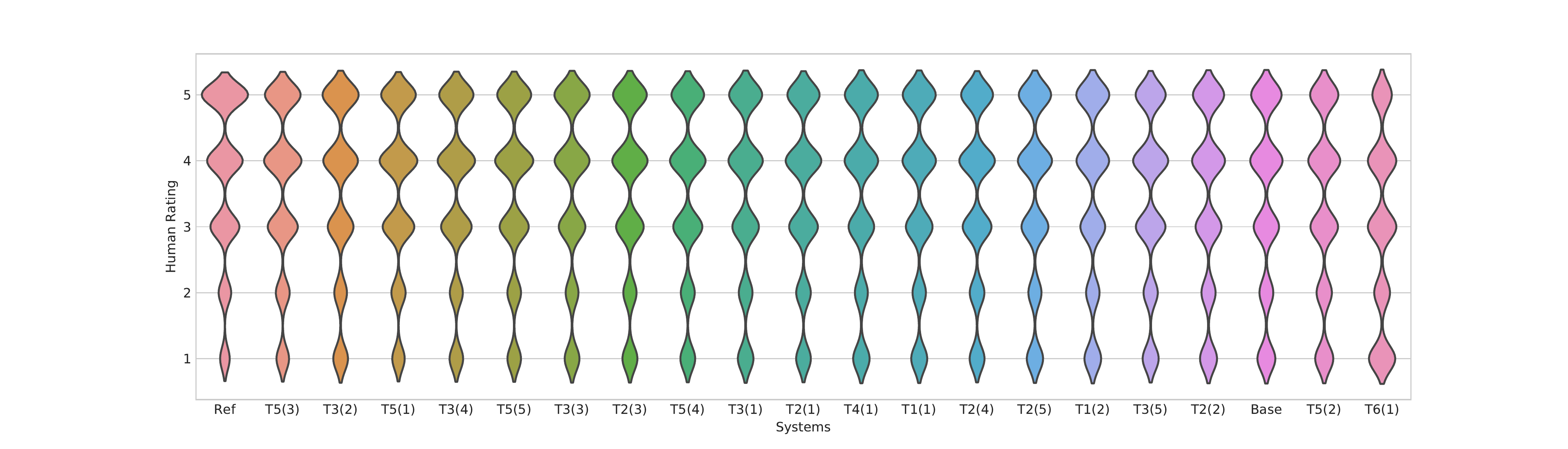}}
\vskip -3mm
\caption{Distribution of human rating score for each level of scores.}
\label{fig:rating-violin2}
\end{figure*}

Figures \ref{fig:human-rating}-\ref{fig:rating-violin2} show the human ratings for each system in several ways.  The systems are shown in the same order on the X axis for all three figures.  Figure \ref{fig:human-rating} shows the mean and the standard deviation of the human ratings for each system (across all responses and all raters for that system). Figure \ref{fig:rating-violin1} shows the distributions of the mean human rating score for each sentence for each system.
Figure \ref{fig:rating-violin2} shows the distribution of all human rating scores for each system across all sentences. 
In this Figure, the area for each score of the violinplot shows a count of the number of scores of each level on the Likert scale.
The "Reference" system (at the far left of each figure) is ratings for the sentences extracted from the original (Twitter) dialog data.  
The Reference system had the best human ratings: it had the highest mean rating in Fig.\ref{fig:human-rating}, the highest median sentence rating in Fig.\ref{fig:rating-violin1} and the most sentences rated as level 5 ("Very good") in Fig.4. 
The worst system (at the right) had a much lower mean rating, and a long tail of poorly rated sentences. 
Table \ref{tab:example} shows examples of the system outputs. 
The humans prefer more supportive responses with sympathy.
The system responses are sometimes rated better than the original human responses.

\begin{table*}
\centering
\caption{Example of System output.}
\label{tab:example}
\begin{tabular}{|l|c|p{13.5cm}|}
\multicolumn{3}{l}{Example 1: The worst system does not show sympathy to the user's troubles.} \\
\hline
Context & & \texttt{U: it 's currently raining on me inside of a bus @njtransit } \\
\hline
Best & 4.0 & \texttt{S: we 're sorry to hear this . please submit a report here so we may further assist you . <URL>}\\
\hline
Worst & 1.0 & \texttt{S: good luck !}\\
\hline
Reference & 3.9 & \texttt{S: hi <USER> , would you please provide me with your origin , destination , bus line , 4 digit vehicle \# \& time of travel ?} \\
\hline
\multicolumn{3}{l}{~}\\
\multicolumn{3}{p{17cm}}{Example 2: The best system shows much deeper sympathy to show how they will solve the user's troubles than the reference. The worst system totally misunderstands the negative context for the positive one.} \\
\hline
Context && \texttt{U: flight canceled . stuck . rebooked 8hrs later in a middle seat and then asked to pay more to change it back to originally booked aisle} \\
\hline
Best & 4.5 & \texttt{S: sorry to hear about this , <USER> . could you dm us your confirmation code so we can take a closer look into this for you ? <URL>} \\
\hline
Worst & 1.2 & \texttt{S: we 're happy to hear that , <USER> . we hope to see you again soon !} \\
\hline
Reference & 3.3 & \texttt{S: sorry to hear about your flight woes . did you know you may be eligible for compensation ? visit <URL>} \\
\hline
\multicolumn{3}{l}{~}\\
\multicolumn{3}{p{17cm}}{Example 3:  The best system shows much deeper sympathy using emotional expressions for the user's troubles than the reference.  The worst system totally misunderstands the negative context for the positive one.} \\
\hline
Context && \texttt{U: savannah was the worst stay over the weekend !} \\
\hline
Best & 4.8 & \texttt{S: oh no ! what happened ? we 'd like to look into this for you . please send us a direct message with your reservation number .}\\
\hline
Worst & 1.5 & \texttt{S: that 's what we like to hear , <USER> ! we 'll be sure to pass this on to the team . have a great day !}\\
\hline
Reference & 4.0 & \texttt{S: we are sorry to hear of your experience . please email us at <E-MAIL> so we can assist you w / this matter .}\\
\hline
\multicolumn{3}{l}{~}\\
\multicolumn{3}{p{17cm}}{Example 4: 
It's difficult to compare the best response sentences assuming the information 
from a back-end system with the reference waiting for a response from a database. 
} \\
\hline
Context && \texttt{U: when are the kaws tee shirts gonna restock again online ?}\\
\hline
Best & 4.7 & \texttt{S: hi <USER> , we don 't have a restock date at this time . please keep an eye on our social media pages for updates .} \\
\hline
Worst & 1.3 & \texttt{S: it does not .} \\
\hline
Reference & 3.7 & \texttt{S: we are checking on the restock date for the kaws tee 's rn .} \\
\hline
\multicolumn{3}{l}{~}\\
\multicolumn{3}{l}{Example 5: The benefits for users and company agents who wants to get more information are different.} \\
\hline
Context && \texttt{U: honestly can 't wait to use my lush bath bomb $\heartsuit$} \\
\hline
Best & 4.4 & \texttt{S: hi <USER> , we 're so pleased to hear that you 're happy with your purchase . enjoy your new goodies !} \\
\hline
Worst & 1.4 & \texttt{S: we have great post-workout meals ! give us a try next ! get 40 \% off : <URL> <URL>}\\
\hline
Reference & 3.6 & \texttt{S: nice which bath bomb ? :D} \\
\hline
\multicolumn{3}{l}{~}\\
\multicolumn{3}{l}{Example 6: More options are good for users. } \\
\hline
Context && \texttt{U: is queen sold out ? you website seems unable to tell me if it is . desperate to know ...}\\
        && \texttt{S: hi <USER> , limited tickets still available . thanks} \\
        && \texttt{U: thanks for the response . so , how does one procure them ?} \\
\hline
Best & 4.4 & \texttt{S: hi , you can order them from our website <URL> or by calling <NUMBERS> . thanks ,}\\
\hline
Worst & 1.1 & \texttt{S: hi <USER> ,}\\
\hline
Reference & 3.9 & \texttt{S: you can follow this link to purchase > > > <URL>}\\
\hline
\end{tabular}
\end{table*}

\vspace{-3mm}
\section{Summary}
This article described the end-to-end conversation modeling track
 of the 6th dialog system technology challenges (DSTC6).
We received 19 system outputs from six teams, and evaluated them based on several objective measures and a human-rating-based subjective measure. The final results are summarized in the article.
Objective and subjective measures were not always consistent across systems. Some mismatches between objective and subjective measures have also been reported in \cite{liu2016not}.
Deeper analysis will be necessary on these results 
to enable us to design a better objective function and perform practical automatic evaluation.

\bibliographystyle{IEEEtran}	
\bibliography{refs}

\newpage

\section{Appendix}	
\begin{appendices}

\section{Dialog Data Format}
\label{apndx:data_format}
\subsection{Basic format}
Each dialog consists of two or more utterances, where each line corresponds to an utterance given by user ‘U:’ or system ‘S:’ indicated at the head of each utterance. An empty line indicates a breakpoint of dialogs. The following text is an example of dialog data file.
\begin{verbatim}  
U: hello !
S: how may I help you ?
U: nothing ...
S: have a good day !

U: your delivery timing & info leaves a lot
to be desired . flowers ordered last wk for
delivery yesterday are nowhere to be seen .
S: hello <USER> , i am sorry for the issue
you are experiencing with us . please dm me
so that i can assist you .
\end{verbatim}

\subsection{Evaluation data format}
Evaluation data basically follows the basic format above, but it also contains partial dialogs that end with a system utterance. With this data, the dialog system has to predict the last utterance in each dialog, where the last utterance is considered the reference. Thus, the system can use the utterances before the last utterance as the context, and predict the last utterance, and is evaluated by comparing the predicted result with the reference.
\begin{verbatim}  
U: hello !                  (context)
S: how may I help you ?     (reference)

U: hello !                  (context)
S: how may I help you ?         :
U: nothing ...                  :
S: have a good day !        (reference)
\end{verbatim}

\subsection{System output format}
Dialog systems are expected to read an evaluation data file and output the following file, where the reference is modified to have ``\texttt{S\_REF:}'' header and a system prediction is appended, which starts with ``\texttt{S\_HYP:}''.
\begin{verbatim}   
U: hello !                  (context)
S_REF: how may I help you ? (reference)
S_HYP: hi .                 (system response)
 
U: hello !                  (context)
S: how may I help you ?         :
U: nothing ...                  :     
S_REF: have a good day !    (reference)
S_HYP: have a nice day !    (system response)
\end{verbatim}  
Note that the references are not disclosed in the challenge period, and replaced with ``\texttt{S: \_\_UNDISCLOSED\_\_}'' in the official evaluation data.
The final result submitted to the challenge does not have to include the reference lines starting with ``\texttt{S\_REF:}''. 

\FloatBarrier
\balance
\end{appendices}

\end{document}